\documentclass[11pt]{article}

\usepackage[margin=1.05in]{geometry}
\usepackage{times}
\usepackage[round]{natbib}
\usepackage{hyperref}
\usepackage{url}
\usepackage{booktabs}
\usepackage{amsmath}
\usepackage{amssymb}
\usepackage{amsthm}
\usepackage{graphicx}
\usepackage{authblk}
\usepackage{multirow}

\newtheorem{proposition}{Proposition}
\newtheorem{corollary}{Corollary}
\newcommand{\Msupp}{M_{\mathrm{supp}}}
\newcommand{\Mwt}{M_{\mathrm{wt}}}
\theoremstyle{definition}

\hypersetup{colorlinks=true, linkcolor=black, citecolor=black, urlcolor=blue}

\title{\bf Mode Collapse Is Cheap to Detect:\\[2pt]
A Ground-Truth-Free Pre-Flight Check for Neural Samplers}

\author[1,2]{Jian Xu}
\affil[1]{RIKEN iTHEMS, Wako, Japan}
\affil[2]{RIKEN Center for Advanced Intelligence Project (AIP), Tokyo, Japan}
\affil[ ]{\texttt{jian.xu@riken.jp}}
\date{}

\begin{document}
\maketitle

\begin{abstract}
Neural samplers are trained against an unnormalised target $\tilde\pi=e^{-E}$ with
no samples from $\pi$, which leaves the practitioner with no way to tell whether
an expensive training run has silently dropped part of the target. The
diagnostics in common use are computed from the model's own draws and are
therefore confined to the model's support: we exhibit a sampler whose
self-normalised effective sample size is $0.99$ while it misses $87\%$ of the
target mass. We argue that \emph{detecting} missing mass is a strictly easier
problem than sampling it: detection needs one point per missed basin plus a local
curvature estimate, whereas correction needs the sampler retrained. We turn this
into a pre-flight check that consumes a few percent of the sampler's own training
budget and uses only $E$, $\nabla E$ and $\nabla^2 E$. On Gaussian-mixture,
Many-Well and rotated anisotropic Many-Well targets with exactly computable
ground truth, the check estimates the missing mass to within $10^{-3}$ at $2.7\%$
of training cost, where a tuned annealed SMC reference needs $70$--$280\%$ of
training cost to do worse. It also applies unchanged to a controlled-SDE sampler
that has no tractable density, where ESS and the ELBO cannot be formed at all.
Separating the missing mass into a part due to lost support and a part due to
misallocated weight sharpens what each diagnostic can do: we prove that the
sampler's own normaliser estimate is unbiased for $Z(1-M_{\mathrm{supp}})$, so
comparing normalising constants -- the reference this literature is measured
against -- recovers the first part exactly and is \emph{identically blind} to the
second at any budget. Across two targets and twelve constructed samplers the SMC
reference duly returns $0.0000$ while up to $64\%$ of the mass is misallocated,
and two runs of the same controlled-SDE architecture on the same target realise
the two failures separately, with nearly equal missing mass and opposite
composition. The estimator carries a \emph{self-diagnostic} that,
without ground truth, is conservative in the safe direction: across $60$
configurations it clears $16$, of which $15$ are accurate to $10^{-2}$ or better.
We are explicit about what this does and does not license. The check cheaply
produces evidence of missing mass, and sometimes evidence that the search has
stabilised, but it cannot certify a run: its thresholds are heuristic, and a mode
whose attraction probability falls below $1/R$ is invisible to the estimator and
to the diagnostic simultaneously -- which is precisely the mode a practitioner
would most want to hear about. We then map the boundary of the
method on a real physical landscape, LJ-13, and report where it fails and why.
\end{abstract}

\section{Introduction}

A neural sampler is trained to draw from $\pi(x)\propto\tilde\pi(x)=e^{-E(x)}$
given only the ability to evaluate $E$ and its derivatives
\citep{zhang2022pis,albergo2025nets,havens2025adjoint,liu2025asbs,guo2026pdns,%
havens2026flowsampling,castromacias2026cds}.
The defining feature of the setting is the absence of samples from $\pi$, and
that absence propagates into evaluation: the quantities practitioners monitor --
the ELBO, the self-normalised effective sample size (ESS), importance-weight
variance -- are expectations under the model. \citet{blessing2024beyond} state
the consequence plainly: such protocols ``rely on samples from the model,
restricting their evaluation capabilities to the model's support,'' and this
``becomes especially problematic when assessing the ability to mitigate mode
collapse.'' \citet{wu2025rdsmc} report the same gap from the other side, noting
that their method ``relies on oracle metrics for hyperparameter tuning'' and
that automating this ``remains an important future direction.'' A practitioner
who has just spent a large compute budget training a sampler has, at present, no
cheap way to learn whether the result is trustworthy.

This paper makes a narrow claim and tests it hard. The claim is an asymmetry:

\begin{quote}
\emph{Detecting that a sampler is missing mass is cheaper than fixing it.}
Detection requires only one point inside each missed basin together with a local
curvature estimate; correction requires the sampler to be retrained.
\end{quote}

If the asymmetry holds, a pre-flight check should exist -- spend a few percent of
the training budget, and cheaply obtain evidence about whether the run dropped
part of the target. We construct such a check
(\S\ref{sec:method}), equip it with a diagnostic that reports when its own answer
should not be believed (\S\ref{sec:selfcheck}), and evaluate it against a tuned
annealed SMC reference at matched cost (\S\ref{sec:exp}). Because a method of
this kind is only as good as its stated scope, we devote \S\ref{sec:scope} to
finding where it breaks, on the 13-atom Lennard-Jones cluster.

\paragraph{Contributions.}
\begin{itemize}
\item We show that the standard ESS diagnostic is blind to the dangerous form of
mode collapse, with an analytic control: a sampler equal to one exact mixture
component attains $\mathrm{ESS}=0.99$ while missing $87\%$ of the mass
(\S\ref{sec:blind}).
\item We give a threshold-free estimator of the missing mass
$M=\sum_j(p_j-q_j)_+$ built from a multi-start second-order mode search, local
harmonic masses, and a Mahalanobis coverage assignment. It uses no samples from
$\pi$ and no estimate of the normalising constant (\S\ref{sec:method}).
\item We give a self-diagnostic, computable without ground truth, which flags
unreliability but cannot certify reliability. It is needed because the estimator
itself carries no fixed error sign under incomplete enumeration
(Proposition~\ref{prop:twosided}). We measure its residual false-clear rate and
identify the search behaviour that defeats it (\S\ref{sec:selfcheck}).
\item We separate the missing mass into support loss and weight misallocation,
prove that ESS is blind to the first (Proposition~\ref{prop:essdelta}) while
normaliser comparison is blind to the second (Corollary~\ref{cor:tilt}) -- so the
two established diagnostic families have disjoint blind spots -- and exhibit both
failures in one architecture on one target (\S\ref{sec:weightcollapse}).
\item We map the method's boundary on LJ-13: the search recovers the global
minimum exactly, but continuous symmetry makes the naive local mass undefined,
and the enumeration is only sufficient at low temperature (\S\ref{sec:scope}).
\end{itemize}

\section{Background: what the usual diagnostics can and cannot see}
\label{sec:blind}

Let $q$ be a trained sampler with tractable density. Importance weights
$w(x)=\tilde\pi(x)/q(x)$ give the self-normalised ESS rate
$\widehat{\mathrm{ESS}} = (\sum_i w_i)^2 / (n\sum_i w_i^2)$, universally reported
as a health check. Its failure mode is structural rather than statistical.
Suppose $\pi=\sum_k \omega_k\,\mathcal N(\mu_k,\sigma_k^2 I)$ with well-separated
components and $q=\mathcal N(\mu_1,\sigma_1^2 I)$ exactly. On the support of $q$
all other components are negligible, so
$w(x)\approx Z\,\omega_1$ is \emph{constant}: the weights are perfectly behaved
and $\widehat{\mathrm{ESS}}\to 1$, while the missing mass is $1-\omega_1$.

We verify this numerically. With $d=16$, $K=8$ and $\omega_1=0.1289$ the analytic
control gives $\widehat{\mathrm{ESS}}\in\{0.979,0.987,0.996\}$ across runs with a
true missing mass of $0.871$. The same phenomenon occurs for genuinely trained
flows and not only for the contrived control: reverse-KL-trained RealNVP flows
that populate $2$ of $8$ components attain
$\widehat{\mathrm{ESS}}\in\{0.907,0.927,0.965\}$ while missing $78$--$87\%$ of the
mass. ESS is high precisely because $q$ fits a \emph{subset} well; the badly
trained sampler that spreads mass into low-density regions is the one ESS
catches. The dangerous case is the one it does not.

\section{A pre-flight check}
\label{sec:method}

\paragraph{Target quantity.} Coverage is not binary: a sampler may place $1\%$ of
its mass on a mode holding $40\%$ of $\pi$. Let the target decompose into basins
with normalised weights $p_j$ under $\pi$ and $q_j$ under $q$. We estimate
\begin{equation}
M \;=\; \sum_j \big(p_j - q_j\big)_+ ,
\label{eq:M}
\end{equation}
the mass of $\pi$ that $q$ under-weights. It is continuous, threshold-free, and
controls the error of any mode-level expectation. It is useful to split it
according to \emph{how} the mass went missing:
\begin{equation}
M \;=\; \underbrace{\sum_{j:\,q_j=0} p_j}_{\Msupp}
   \;+\; \underbrace{\sum_{j:\,q_j>0}\big(p_j-q_j\big)_+}_{\Mwt} .
\label{eq:decomp}
\end{equation}
$\Msupp$ is collapse in the usual sense -- basins the sampler never enters --
and is what the phrase ``mode collapse'' normally denotes. $\Mwt$ is collapse
\emph{without} support loss: every basin is reached, but the proportions among
them are wrong. The distinction is not cosmetic. Section~\ref{sec:theory} shows
that the two established diagnostic families each see exactly one of the two
terms, and \S\ref{sec:weightcollapse} exhibits two samplers of the same
architecture on the same target with nearly the same $M$ and opposite
decompositions. The estimator defined here is indifferent to the split: it
compares a mass budget against an occupancy and never asks which kind of
discrepancy produced the gap.

\begin{figure}[t]
\centering
\includegraphics[width=\textwidth]{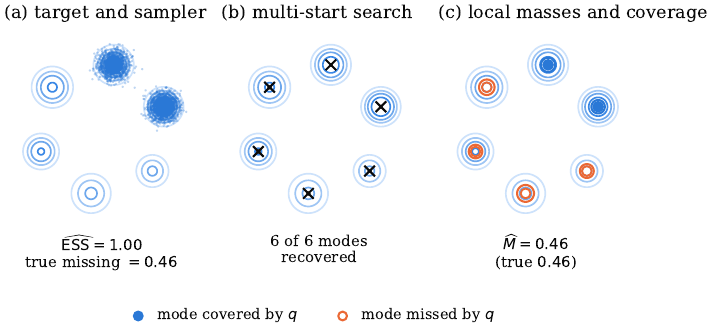}
\caption{The check in two dimensions, on the configuration of
Proposition~\ref{prop:ess}. \textbf{(a)}~A six-component target (contours) and a
sampler that reproduces two of the components exactly and places no mass
elsewhere (dots). Its self-normalised ESS is $1.00$ -- the largest value the
diagnostic can take -- while $46\%$ of the target mass is absent.
\textbf{(b)}~Multi-start second-order search from a broad initialisation recovers
all six modes, including the four the sampler never visits.
\textbf{(c)}~Laplace ellipses at the recovered modes, filled where the sampler's
draws populate them and open where they do not, giving
$\widehat M=0.46$ against a true $0.46$. Nothing in the pipeline uses samples
from $\pi$.}
\label{fig:concept}
\end{figure}

\paragraph{Step 1: mode search.} We enumerate candidate maxima of $\log\tilde\pi$
by multi-start optimisation from a broad initialisation, using Adam with cosine
decay followed by Levenberg--Marquardt Newton refinement. The second-order stage
is not a refinement of convenience. On a quartic target a fixed-step ascent
diverges outright (gradients reach $\sim\!10^3$ at a broad initialisation), and
on an anisotropic target Adam alone leaves siblings of the \emph{same} basin
$\sim\!18$ units apart, many at saddle points with negative Hessian eigenvalues,
which destroys both the deduplication and the coverage test downstream. With LM
Newton refinement the search recovers exactly $32$ of $32$ modes on our hardest
synthetic target.

\paragraph{Step 2: local masses.} At each converged candidate $x_j^\star$ with
$A_j=-\nabla^2\log\tilde\pi(x_j^\star)\succ0$ we take the Laplace mass
\begin{equation}
\log m_j \;=\; \log\tilde\pi(x_j^\star) + \tfrac{d}{2}\log 2\pi
- \tfrac12\log\det A_j ,
\qquad
\hat p_j \;=\; \frac{m_j}{\sum_k m_k}.
\label{eq:laplace}
\end{equation}
Duplicates are merged by Mahalanobis distance under $A_j$; a Euclidean tolerance
is not scale-invariant and fails as soon as modes are anisotropic. Note that
\eqref{eq:M} is a \emph{ratio} of harmonic masses, so a systematic Laplace bias
shared across modes cancels -- which is why accuracy on anharmonic targets is
better than a naive error analysis would suggest (\S\ref{sec:exp}).

\paragraph{Step 3: coverage.} Draws from $q$ cost no target evaluations. We draw
$n_q$ of them and assign each to the nearest discovered mode in the local
Mahalanobis metric, mirroring how the oracle assigns them, giving $\hat q_j$ with
$\sum_j\hat q_j=1$. A single-point test on $\log q(x_j^\star)$ was our first
choice and is not robust: under anisotropy it rejects modes that $q$ does cover.

\paragraph{What is deliberately absent.} An earlier version normalised by
$\hat Z_q$, the self-normalised estimate of the covered mass. For a collapsed $q$
the weights are heavy-tailed and $\hat Z_q$ swung enough to move the reported
answer from $0.55$ to $0.38$ while the reference $\log Z$ was essentially exact.
Equations \eqref{eq:M}--\eqref{eq:laplace} never form $\hat Z_q$.

\paragraph{What \eqref{eq:M} estimates, and what it does not.} Both $\hat p$ and
$\hat q$ are normalised over the \emph{discovered} modes: $\hat p_j=p_j/P_S$ with
$P_S=\sum_{k\in S}p_k$, and the draws of $q$ that really belong to an
undiscovered basin are assigned to their nearest discovered neighbour. Under
complete enumeration $P_S=1$ and $\widehat M$ is consistent for $M$
(Proposition~\ref{prop:bias} controls the remaining Laplace error). Under
\emph{incomplete} enumeration both quantities are inflated and the error is
\textbf{not} signed in general -- Proposition~\ref{prop:twosided} exhibits
incomplete-enumeration instances erring in each direction. We therefore make no
bound claim: $\widehat M$ estimates the missing mass of the landscape the search
actually resolved, and the diagnostic of \S\ref{sec:selfcheck} exists to detect
when that landscape is not the whole one.

\section{What the estimator can and cannot guarantee}
\label{sec:theory}

Four short results delimit the method. The first formalises the failure of ESS
that motivates the paper; the second says no method of this kind can ever
certify its answer; the third explains why an approximation as crude as Laplace
suffices; the fourth locates the blind spot of the self-diagnostic. Proofs are
given in Appendix~\ref{app:proofs}.

\begin{proposition}[Subset collapse is invisible to ESS]
\label{prop:ess}
Let $S$ be a union of basins with $\pi(S)=\omega\in(0,1)$ and let
$q(x)=\pi(x)\mathbf 1_S(x)/\omega$, i.e.\ $q$ reproduces $\pi$ exactly on $S$ and
places no mass elsewhere. Then the importance weights are constant on
$\operatorname{supp}q$, so the self-normalised $\widehat{\mathrm{ESS}}=1$ for
every sample size, while the missing mass is $M=1-\omega$, which can be made
arbitrarily close to $1$.
\end{proposition}

The point is not that ESS is noisy but that it is \emph{exactly} maximal in the
worst case. The idealisation is not needed: if $q$ is merely \emph{close} to
$\pi(\cdot\mid S)$ the conclusion survives with a constant.

\begin{proposition}[ESS is bounded below independently of the missing mass]
\label{prop:essdelta}
Let $q$ be supported on $S$ and $\delta$-accurate there, i.e.\
$\big|\log\{q(x)/\pi(x\mid S)\}\big|\le\delta$ for all $x\in S$. Then the
population ESS rate satisfies $\mathrm{ESS}(q)\ge e^{-4\delta}$, a bound that
does not involve $\Msupp$.
\end{proposition}

Since $\delta$ and $\Msupp$ vary independently, ESS carries no information about
how much mass is absent: it measures how well $q$ fits whatever it does cover.
Our measurements ($\widehat{\mathrm{ESS}}=0.99$ at $M=0.87$) are the
finite-sample shadow of this identity, and the spread we observe across runs
($0.979$--$0.996$) is the $\delta>0$ correction.

Weight collapse is the opposite case. If $q$ reaches every basin but with the
wrong proportions, then $w\propto p_j/q_j$ varies across basins by construction,
so the weights are not constant and ESS \emph{does} degrade. ESS is not a broken
diagnostic; it is matched to one of the two terms in \eqref{eq:decomp} and blind
to the other. The following results show that the natural competitor to our
check has precisely the complementary blind spot.

\begin{proposition}[Normalizer identity]
\label{prop:identity}
Let $q$ be supported on $S$, a union of basins. Then
\begin{equation}
\mathbb E_q\!\left[\tilde\pi/q\right] \;=\; \int_S \tilde\pi \;=\; Z\,\pi(S)
\;=\; Z\,(1-\Msupp),
\qquad\text{so}\qquad \Msupp \;=\; 1 - \mathbb E_q[\tilde\pi/q]\,/\,Z .
\label{eq:identity}
\end{equation}
\end{proposition}

The importance-sampling estimate of the normalising constant under a collapsed
sampler is unbiased not for $Z$ but for $Z(1-\Msupp)$. This is the precise sense
in which comparing normalising constants is the \emph{right} competitor for our
check -- the missing mass is exactly the gap -- and it explains the empirical
behaviour of the SMC reference in \S\ref{sec:exp}. It also has two consequences
that bound what that competitor can do.

\begin{corollary}[Blindness to weight collapse]
\label{cor:tilt}
If $\operatorname{supp}(q)\supseteq\operatorname{supp}(\pi)$ then
$\mathbb E_q[\tilde\pi/q]=Z$ exactly, however badly $q$ misallocates mass among
basins. Hence any diagnostic of the form
$\widehat M = 1-\widehat Z_q/\widehat Z_{\mathrm{ref}}$ has population value $0$
while $\Mwt$ may be as large as $1-\max_j p_j$.
\end{corollary}

\begin{corollary}[The failure direction is a false negative]
\label{cor:silent}
The same estimator returns $\widehat M\le 0$ whenever
$\log\widehat Z_{\mathrm{ref}}\le\log\widehat Z_q$. As $\widehat Z_q$ is computed
under a concentrated $q$ while $\widehat Z_{\mathrm{ref}}$ must integrate the
whole target, this is the typical situation at small reference budgets.
\end{corollary}

Corollary~\ref{cor:tilt} is the sharper statement. It is not about estimator
variance, tuning, or budget: the estimand itself does not depend on $\Mwt$, so no
amount of computation makes a normalizer comparison sensitive to weight collapse.
Corollary~\ref{cor:silent} explains the failure mode we measure in
\S\ref{sec:exp}, where the reference returns exactly $0.0000$ -- ``nothing is
missing'' -- at budgets where its $\log Z$ estimate is still far off. Our own
estimator sidesteps both: \eqref{eq:M}--\eqref{eq:laplace} never form
$\widehat Z_q$, a choice we originally made for variance reasons
(\S\ref{sec:method}) and which Corollary~\ref{cor:tilt} now justifies on
identifiability grounds.

\begin{proposition}[No certificate from finitely many queries]
\label{prop:nocert}
Fix any algorithm that outputs an estimate of $M$ after $N$ evaluations of
$\tilde\pi,\nabla\tilde\pi,\nabla^2\tilde\pi$ at adaptively chosen points, and
which reads those values at finite precision $\varepsilon>0$. For every $N$,
every $\varepsilon>0$ and every $\delta>0$ there exist two targets
$\tilde\pi_1,\tilde\pi_2$ whose values and first two derivatives differ by less
than $\varepsilon$ at all $N$ queried points -- so the algorithm returns the same
answer on both -- and for which $|M(\tilde\pi_1)-M(\tilde\pi_2)|\ge 1-\delta$.
It suffices to separate the two targets by a unit-variance Gaussian bump placed
at distance
\begin{equation}
r \;\ge\; \max\Big\{4,\; 2\sqrt{\log\!\big(\lambda\,(2\pi)^{-d/2}/\varepsilon\big)}\Big\},
\qquad \lambda \;=\; Z_1(1-\delta)/\delta,
\label{eq:bumpr}
\end{equation}
from every queried point, which is possible inside any ball of radius exceeding
$r\,N^{1/d}$. Consequently, finite-query evidence can
certify only mass associated with regions the queries actually reached; it
cannot certify the absence of additional unseen mass.
\end{proposition}

This says that \emph{no} method of this family can certify its answer, and it is
why \S\ref{sec:selfcheck} reports a necessary rather than a sufficient condition.
It does \emph{not} say that our particular estimator errs in a known direction --
that is a separate question, settled negatively by
Proposition~\ref{prop:twosided}. It is the same obstruction that makes
simulation-free mode discovery hard \citep{he2025notrick}: the two statements are
duals, one about finding mass and one about certifying its absence.

\begin{proposition}[Incomplete enumeration is not one-sided]
\label{prop:twosided}
With $S$ the set of discovered modes, $\widehat M$ can exceed or fall below $M$.
\emph{Over-estimation:} take $p=(0.4,0.3,0.3)$, $q=(0.2,0.4,0.4)$, so $M=0.2$;
if the search finds only the first two modes and the third mode's $q$-draws are
assigned to the second, then $\hat p=(0.571,0.429)$, $\hat q=(0.2,0.8)$ and
$\widehat M=0.371>M$. \emph{Under-estimation:} take two modes of weight $0.5$
with $q$ concentrated entirely on the first, so $M=0.5$; if the search finds only
the first, $\hat p=\hat q=1$ and $\widehat M=0<M$.
\end{proposition}

The mechanism is the shared normalisation: dividing by $P_S<1$ inflates $\hat p$,
while re-assigning orphaned draws inflates $\hat q$, and which effect dominates
depends on where the undiscovered mass sits relative to the sampler. This is the
precise sense in which the estimate is conditional on the search, and it is why
we report the diagnostic of \S\ref{sec:selfcheck} alongside every number rather
than claiming a bound.

\begin{proposition}[Only the \emph{spread} of the Laplace bias matters]
\label{prop:bias}
Suppose the local mass of every discovered mode is estimated with a
multiplicative error $\hat m_j=c_j m_j$ where $c_j\in[c(1-\eta),\,c(1+\eta)]$ for
some common $c>0$ and $\eta\in[0,1)$. Then the estimator satisfies
\begin{equation}
\big|\widehat M - M\big| \;\le\; \frac{2\eta}{1-\eta}.
\end{equation}
In particular, if the bias is the same across modes ($\eta=0$) the estimator is
exact, however large that bias is.
\end{proposition}

This explains an otherwise surprising measurement. On Many-Well every mode is a
copy of the same quartic well, so the Laplace bias is shared and $\eta\approx0$;
the estimator reaches error $10^{-3}$ on a target whose modes are manifestly not
Gaussian. Accuracy is governed by how \emph{uniform} the anharmonicity is across
modes, not by how severe it is.

\begin{proposition}[Blind spot of the growth statistic]
\label{prop:blind}
Model the search as $R$ i.i.d.\ restarts, and let mode $j$ have probability
$\alpha_j$ of being reached by one restart. If mode $j$ is missed at budget $R$
and at the reference budget $R/4$, it contributes nothing to either the weight
ratio or the count ratio, and the configuration is cleared as though it had been
enumerated. This happens with probability $(1-\alpha_j)^{R}$, so for
$\alpha_j\lesssim 1/R$ a mode carrying arbitrary weight can be missed by the
estimator \emph{and} by the diagnostic simultaneously.
\end{proposition}

The proposition predicts precisely the failure we observe: at $R=32$ on the
Gaussian mixture, four of eight modes have small attraction volume, both budgets
return the same four, the ratios are $1.000$, and the estimate is wrong by
$0.29$. It also says what to do about it -- the failure probability decays
geometrically in $R$, which is why the cleared verdicts at $R=1024$ are reliable
in every run.

\section{Knowing when not to believe the answer}
\label{sec:selfcheck}

A diagnostic that cannot fail loudly is worse than none. We report two statistics,
neither of which uses ground truth.

\textbf{(a) Growth test.} Let $W(R)$ be the total harmonic mass of the modes
discovered by $R$ restarts, and $N(R)$ the number of them. We report
$\rho = W(R)/W(R/4)$ together with the count ratio $N(R)/N(R/4)$. If the
enumeration has saturated both are $\approx1$.

\textbf{(b) Singleton mass.} The fraction of discovered mass held by modes that
exactly one trajectory reached, in the spirit of Good--Turing.

We declare the estimate reliable when $\rho<1.1$, the count ratio is below $1.1$,
and the singleton fraction is below $0.05$. These three numbers are heuristic.
They were not calibrated on a held-out family of targets, and we do not claim
they transfer; what the experiments below establish is that \emph{some} threshold
in this family separates the accurate from the inaccurate configurations on our
benchmarks, not that these particular values are the right ones elsewhere.
Calibrating them -- ideally to a target false-clear rate -- is the obvious next
step and we have not taken it.

\paragraph{Why the count ratio is not redundant.} Weight growth alone is
insufficient, and its failure is instructive. When basins have very unequal
\emph{attraction} volumes, a small search returns the same easy subset at both
budgets: $\rho=1.000$, no singletons, and yet half the modes have never been
visited. We observed exactly this -- $4$ of $8$ modes found, $\rho=1.000$,
estimate off by $0.29$. Discovered weight is blind to it; the discovered
\emph{count} is not. Adding the count ratio removed three of the four such
failures in our sweep.

\paragraph{The diagnostic is one-sided.} This is not a threshold to be tuned away.
A search that never reaches a basin at any budget in the tested range produces no
signal at all, in either statistic. The check is therefore a necessary and not a
sufficient condition: $\rho>1.1$ means do not trust the estimate; $\rho<1.1$ means
only that the search found no evidence against it. Stated as a capability, the
method cheaply produces evidence of missing mass, and sometimes evidence that the
search has stabilised; it does not license the stronger reading that a cleared
run may be trusted. That is the same asymmetry the estimator
itself carries (\S\ref{sec:method}), and it is inherited from the same source.

\begin{figure}[t]
\centering
\includegraphics[width=\textwidth]{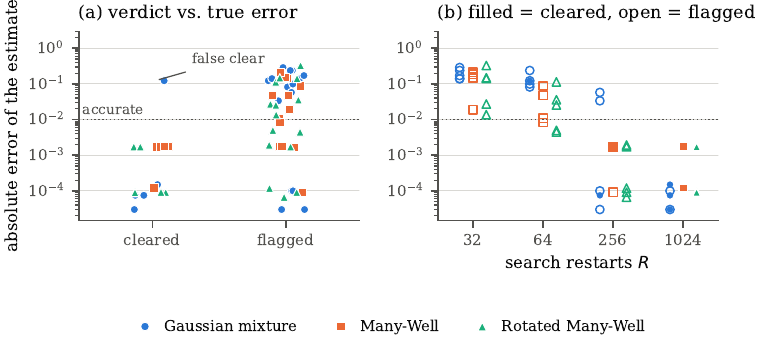}
\caption{The self-diagnostic against ground truth it never sees, over all $60$
configurations. \textbf{(a)}~Error of the estimate, grouped by verdict; the
dotted line marks $10^{-2}$. Of $16$ cleared configurations, $15$ are accurate.
\textbf{(b)}~The same points against the search budget; filled markers are
declared reliable. Both the error and the verdict improve with $R$, which is what
makes ``increase the budget'' the right response to a warning.}
\label{fig:selfcheck}
\end{figure}

\paragraph{What it does in practice.} Figure~\ref{fig:selfcheck} shows all $60$
configurations (three targets, four search budgets, five seeds). The check is
conservative: it clears $16$ of $60$, and $15$ of those $16$ have error below
$10^{-2}$ -- most of them below $10^{-3}$. The single false clear sits at the
smallest useful budget on the Gaussian mixture. Many accurate configurations are
also flagged, which is the harmless direction of the trade: the actionable output
is ``increase the search budget'', and panel~(b) shows why that instruction is
sound -- error falls and verdicts flip to reliable as $R$ grows.

\begin{table}[t]
\caption{Self-diagnostic by target and search budget, five seeds each. The verdict
column counts how many of the five seeds were cleared.}
\label{tab:selfcheck}
\centering
\begin{tabular}{lrrr}
\toprule
target & restarts $R$ & mean abs.\ error & cleared \\
\midrule
Gaussian mixture & 32   & 0.1955 & 0/5 \\
Gaussian mixture & 64   & 0.1330 & 1/5 \\
Gaussian mixture & 256  & 0.0183 & 1/5 \\
Gaussian mixture & 1024 & 0.0001 & 3/5 \\
\midrule
Many-Well & 32   & 0.1434 & 0/5 \\
Many-Well & 64   & 0.0403 & 0/5 \\
Many-Well & 256  & 0.0014 & 1/5 \\
Many-Well & 1024 & 0.0014 & 5/5 \\
\midrule
Rotated Many-Well & 32   & 0.1318 & 0/5 \\
Rotated Many-Well & 64   & 0.0370 & 0/5 \\
Rotated Many-Well & 256  & 0.0008 & 0/5 \\
Rotated Many-Well & 1024 & 0.0007 & 5/5 \\
\bottomrule
\end{tabular}
\end{table}

\section{Experiments}
\label{sec:exp}

\paragraph{Targets with exact ground truth.} We deliberately avoid targets whose
``gold'' is itself an expensive approximation, since the gold would be the object
under suspicion. All three targets admit exact ground truth. \textbf{(i)}~A
Gaussian mixture, $d=16$, $K=8$: $Z$ and the mode weights are analytic.
\textbf{(ii)}~\textbf{Many-Well}, $E=\sum_{i<m}(x_i^4-6x_i^2-\tfrac12 x_i)+\sum_{i\ge m}\tfrac12 x_i^2$
with $d=16$, $m=5$, giving $2^5=32$ anharmonic and asymmetric modes; the energy
separates over coordinates, so $\log Z$ and every sign-pattern weight reduce to
one-dimensional quadrature. \textbf{(iii)}~\textbf{Rotated Many-Well},
$E'(x)=E(Ax)$ with $A=Q\,\mathrm{diag}(s)$, $Q$ random orthogonal and condition
number $20$: this removes coordinate separability, isotropy and Hessian
conditioning simultaneously, while $\log Z' = \log Z - \log|\det A|$ stays exact.

\paragraph{Why this reference, and what it cannot settle.} We compare against an
annealed SMC estimate of $\log Z$, with the missing mass read off the gap to the
sampler's own estimate. Proposition~\ref{prop:identity} is what justifies the
choice: the gap \emph{is} the missing mass due to lost support, so this is the
statistic a careful practitioner would reach for, not an invented weakling.
But the same analysis makes the comparison partly circular, and we say so
rather than let the reader discover it: Corollary~\ref{cor:tilt} shows the
reference is blind to $\Mwt$ at the population level, so beating it on weight
collapse (\S\ref{sec:weightcollapse}) is guaranteed in advance and demonstrates a
property of the estimand, not a property of our search. Only the $\Msupp$
comparison below is a genuine contest. \S\ref{sec:altsearch} adds a competitor
that estimates basin weights directly and is therefore not subject to
Corollary~\ref{cor:tilt} at all.

\paragraph{Reference.} The competitor is not a straw man. A weak annealed
importance sampler (fixed step, one MALA move per level, no resampling) fails so
badly here -- $\log Z$ off by $4.1$ nats, never firing at all -- that it would
have suggested a $200\times$ advantage. We therefore use a full SMC sampler:
geometric annealing, several MALA moves per level with dual-averaging step
adaptation to $0.574$ acceptance, and ESS-triggered systematic resampling. The
cost accounting is generous to it, charging two target evaluations per particle
per MALA move. Upgrading the reference reduced our measured advantage by an order
of magnitude, and we report the post-upgrade numbers.

\begin{figure}[t]
\centering
\includegraphics[width=\textwidth]{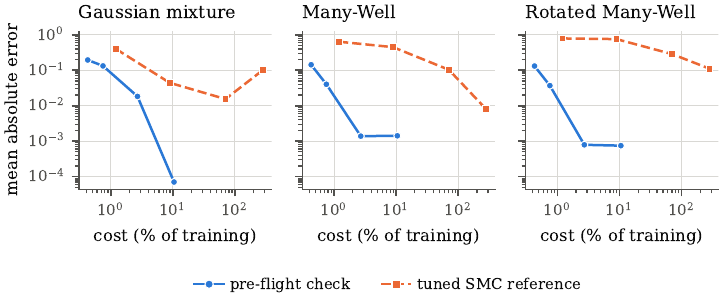}
\caption{Error--cost frontier, five seeds. The check (blue) sits below and to the
left of the tuned SMC reference (orange) on every target, and the gap widens as
the target gets harder: on the rotated anisotropic target the reference has not
become reliable even at $2.8\times$ the sampler's training cost.}
\label{fig:frontier}
\end{figure}

\paragraph{Results.} Table~\ref{tab:main} and Figure~\ref{fig:frontier} give the
mean absolute error over five seeds against the best SMC configuration on each
target. On the two harder
targets the check is both cheaper and more accurate by two orders of magnitude.
On the Gaussian mixture the reference is marginally more accurate at $R{=}256$,
at $26\times$ the cost; raising the search budget to $R{=}1024$ ($10.4\%$ of
training) reduces the error to $0.0001$, and -- importantly -- this is exactly
the configuration change that the self-diagnostic of \S\ref{sec:selfcheck} calls
for, without being told the answer (Table~\ref{tab:selfcheck}).

\begin{table}[t]
\caption{Mean absolute error of the estimated missing mass over five seeds,
against the best configuration of a tuned SMC reference. Cost is a fraction of
the sampler's own training budget.}
\label{tab:main}
\centering
\begin{tabular}{lrrrr}
\toprule
& \multicolumn{2}{c}{pre-flight check} & \multicolumn{2}{c}{best SMC reference} \\
\cmidrule(lr){2-3}\cmidrule(lr){4-5}
target & cost & error & cost & error \\
\midrule
Gaussian mixture  & 2.7\%  & 0.0183 & 70.5\%  & 0.0154 \\
Gaussian mixture  & 10.4\% & 0.0001 & 70.5\%  & 0.0154 \\
Many-Well         & 2.7\%  & 0.0014 & 281.7\% & 0.0080 \\
Rotated Many-Well & 2.7\%  & 0.0008 & 281.7\% & 0.1108 \\
\bottomrule
\end{tabular}
\end{table}

\paragraph{How the cost scales with dimension.} The headline figure of a few
percent of training cost is dimension-specific, and we measure how it degrades.
Table~\ref{tab:dim} runs the check on Many-Well at $d=16$ to $128$ with the
sampler and search budget fixed, reporting both target-evaluation counts and
wall-clock. Two things happen. The accuracy is essentially flat, staying
at $1$--$2\times10^{-3}$ across the whole range, because the number of modes and
the quality of the harmonic approximation are unchanged. The cost does move, close to linearly in
$d$, because assembling a full Hessian costs $O(d)$ backward passes. Wall-clock
grows faster than the evaluation count ($18.7\%$ versus $11.6\%$ at $d=128$),
which is the concrete sense in which a Hessian evaluation and a gradient
evaluation are not interchangeable units, and a reason to prefer the wall-clock
column when judging the method.

Extrapolating the observed linear trend, the check reaches parity with the
training budget at $d\approx700$ measured by wall-clock, or $d\approx10^{3}$
measured by evaluation count, and is more expensive than retraining beyond
that. We therefore state the claim precisely: on targets of the size used in the
current sampling benchmarks the check costs a few percent of training, and this
is not a statement about $d=10^{3}$--$10^{4}$ problems. The standard remedy --
replacing the explicit Hessian by Hessian-vector products with a Lanczos or
stochastic-trace estimator of $\log\det$, which would make the per-mode cost
$O(k)$ rather than $O(d)$ -- applies directly here, but we have not implemented
it and do not claim it.

\begin{table}[t]
\caption{Cost of the check as the dimension grows, Many-Well with $m=5$
($32$ modes), $R=256$, three seeds. Accuracy is flat; cost grows about linearly
in $d$ because a full Hessian needs $O(d)$ backward passes.}
\label{tab:dim}
\centering
\begin{tabular}{rrrr}
\toprule
$d$ & target evals (\% of training) & wall-clock (\% of training) & abs.\ error \\
\midrule
16  & 2.69\%  & 2.7\%  & 0.0012 \\
32  & 3.96\%  & 4.4\%  & 0.0017 \\
64  & 6.49\%  & 9.2\%  & 0.0017 \\
128 & 11.57\% & 18.7\% & 0.0017 \\
\bottomrule
\end{tabular}
\end{table}

\paragraph{A sampler with no density at all.} Every experiment above used a
normalizing flow, which has a tractable $q$ and therefore an ESS -- so a sceptic
may object that the check is only interesting because we chose to ignore a
diagnostic that existed. We therefore train a Path-Integral-Sampler style
controlled SDE \citep{zhang2022pis} on Many-Well: it emits samples and nothing
else, so $q(x)$, the ESS, the ELBO and importance weights are all undefined, and
no diagnostic in current use applies. The check is unaffected, because its
coverage step consumes draws from $q$ and never its density. Across five seeds
the trained sampler populates $1$ of $32$ modes with true missing mass
$0.569$, and the check recovers it to a mean absolute error of $0.0027$ at
$R{=}256$ ($0.55\%$ of training cost) and $0.0025$ at $R{=}1024$ ($2.21\%$),
with a per-seed spread of $0.0017$--$0.0033$. The self-diagnostic clears
$R{=}1024$ in all five runs and flags $R{=}256$ in four of five, again erring
conservatively. This is the setting the method was built for: the sampler is
opaque, the target is not, and everything the check needs comes from the
target.

\paragraph{The error decomposes.} Total error splits into search incompleteness
and Laplace bias, and the second term is negligible because \eqref{eq:M} is a
ratio: on Many-Well, where every mode has the same quartic shape, the bias
cancels almost exactly and the error is $0.001$ despite the modes being
manifestly non-Gaussian. What remains is search incompleteness, controlled by a
single interpretable knob -- the number of restarts -- and reported by the
self-diagnostic. The SMC reference, by contrast, never becomes reliable on the
rotated target: its mean error is $0.287$ at $70.5\%$ of training cost and
$0.111$ at $281.7\%$, both far above what the check reaches for $2.7\%$.

\section{Collapse without support loss}
\label{sec:weightcollapse}

Everything so far concerned $\Msupp$: samplers that drop basins. This section
measures the other term. It matters for two reasons. Corollary~\ref{cor:tilt}
says the SMC reference of \S\ref{sec:exp} cannot see $\Mwt$ at any budget, so if
$\Mwt$ occurs in practice the competitor is not merely expensive but inapplicable;
and $\Mwt$ lives at small $M$, a regime the experiments above never entered.

\paragraph{Constructed collapse, exactly decomposed.} We build samplers
whose $(\Msupp,\Mwt)$ are known exactly. A \emph{tilted} sampler assigns basin
$j$ probability $v_j\propto p_j^{\tau}$ and is exact \emph{within} every basin,
so $\operatorname{supp}(q)=\operatorname{supp}(\pi)$, $\Msupp=0$, and $M=\Mwt$ is
tuned by $\tau$. A \emph{mixed} sampler additionally drops the lightest quarter
of the basins, making both terms nonzero. On the Gaussian mixture this is a
mixture over a component subset; on Many-Well a basin is a sign pattern over the
first $m$ coordinates and we draw each coordinate from its exact within-well
conditional by inverse CDF, so only the basin weights are wrong.

\begin{table}[t]
\caption{Constructed collapse with exactly known decomposition. In the tilt
family $\Msupp=0$, so by Corollary~\ref{cor:tilt} the normalizer gap must be zero
whatever $\Mwt$ is: it is, and the SMC estimate is exactly $0.0000$ in all twelve
tilt rows. The mixed rows vary $\Mwt$ at fixed $\Msupp$ and are the cleaner
demonstration -- on the GMM the gap reads $0.104$ three times while $\Mwt$ grows
from $0$ to $0.059$. ESS, blind to $\Msupp$, does respond here. Search budget
$R{=}1024$ on the GMM and $R{=}256$ on Many-Well, three repetitions.}
\label{tab:weightcollapse}
\begin{center}
\small
\begin{tabular}{llccccccc}
\toprule
target & family & $\Msupp$ & $\Mwt$ & $M$ & ESS & norm.\ gap & SMC & ours \\
\midrule
\multirow{9}{*}{GMM}
 & tilt $\tau{=}1$    & 0 & 0.000 & 0.000 & 1.000 & $-0.000$ & 0.000 & 0.006 \\
 & tilt $\tau{=}0.5$  & 0 & 0.083 & 0.083 & 0.959 & $+0.001$ & 0.000 & 0.086 \\
 & tilt $\tau{=}0.25$ & 0 & 0.128 & 0.128 & 0.909 & $+0.002$ & 0.000 & 0.131 \\
 & tilt $\tau{=}0$    & 0 & 0.180 & 0.180 & 0.843 & $+0.003$ & 0.000 & 0.179 \\
 & tilt $\tau{=}2$    & 0 & 0.137 & 0.137 & 0.843 & $-0.002$ & 0.000 & 0.138 \\
 & tilt $\tau{=}3$    & 0 & 0.235 & 0.235 & 0.498 & $-0.004$ & 0.000 & 0.236 \\
 & mixed $\tau{=}1$   & 0.104 & 0.000 & 0.104 & 1.000 & $+0.104$ & 0.058 & 0.104 \\
 & mixed $\tau{=}0.5$ & 0.104 & 0.010 & 0.114 & 0.984 & $+0.104$ & 0.060 & 0.116 \\
 & mixed $\tau{=}0$   & 0.104 & 0.059 & 0.162 & 0.937 & $+0.104$ & 0.064 & 0.165 \\
\midrule
\multirow{9}{*}{Many-Well}
 & tilt $\tau{=}1$    & 0 & 0.000 & 0.000 & 1.000 & $-0.000$ & 0.000 & 0.007 \\
 & tilt $\tau{=}0.5$  & 0 & 0.297 & 0.297 & 0.621 & $+0.003$ & 0.000 & 0.299 \\
 & tilt $\tau{=}0.25$ & 0 & 0.481 & 0.481 & 0.338 & $+0.008$ & 0.000 & 0.482 \\
 & tilt $\tau{=}0$    & 0 & 0.637 & 0.637 & 0.144 & $+0.005$ & 0.000 & 0.639 \\
 & tilt $\tau{=}2$    & 0 & 0.417 & 0.417 & 0.160 & $-0.010$ & 0.000 & 0.416 \\
 & tilt $\tau{=}3$    & 0 & 0.540 & 0.540 & 0.013 & $-0.037$ & 0.000 & 0.539 \\
 & mixed $\tau{=}1$   & 0.008 & 0.000 & 0.008 & 1.000 & $+0.008$ & 0.000 & 0.012 \\
 & mixed $\tau{=}0.5$ & 0.008 & 0.265 & 0.273 & 0.651 & $+0.003$ & 0.000 & 0.275 \\
 & mixed $\tau{=}0$   & 0.008 & 0.575 & 0.583 & 0.190 & $-0.001$ & 0.000 & 0.585 \\
\bottomrule
\end{tabular}
\end{center}
\end{table}

\begin{figure}[t]
\begin{center}
\includegraphics[width=0.99\textwidth]{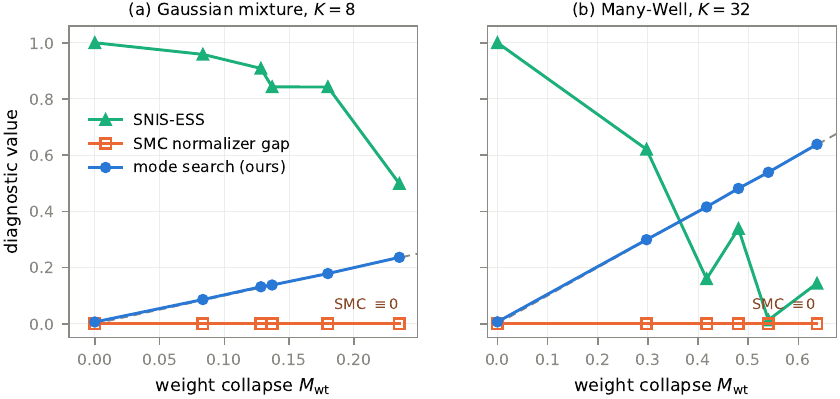}
\end{center}
\caption{The blind spot of the normalizer comparison. Within each panel every
sampler reaches every basin ($\Msupp=0$) and only the proportions are wrong. The
SMC estimate built on comparing normalising constants stays pinned at zero while
more than half the target mass is misallocated, as Corollary~\ref{cor:tilt}
requires. ESS responds; the check tracks the truth.}
\label{fig:blindspot}
\end{figure}

Table~\ref{tab:weightcollapse} and Figure~\ref{fig:blindspot} give the result.
Across the twelve tilt rows $\Mwt$ ranges over $[0,0.637]$ while the normalizer
gap stays within $0.037$ of zero and the SMC estimate is exactly $0.0000$ in
every one. The mixed rows isolate the mechanism: on the GMM the gap reads
$0.1038$, $0.1041$, $0.1043$ against a true $\Msupp$ of $0.1038$ while $\Mwt$
grows from $0$ to $0.059$. The gap is measuring $\Msupp$ and nothing else, to
three decimals, exactly as \eqref{eq:identity} predicts. The check tracks $M$
throughout, with absolute error at most $0.007$ over all eighteen rows and at
most $0.003$ on the sixteen with $M>0$.

\paragraph{The same architecture collapses in both ways.} The controlled-SDE
audit of \S\ref{sec:exp} produced a sampler that populates $1$ of $32$ basins,
i.e.\ $M$ is almost entirely $\Msupp$. Retraining the same PIS-style control on
the same Many-Well target under a different gradient-clipping setting inside the
control produces a sampler that reaches \emph{all} $32$ basins and is wrong about
the proportions instead. We report this as an existence result and do not claim
clipping as the general cause: in our own paired runs it is the only setting that
differs, but we did not sweep it. The two failures have nearly the same total
missing mass and opposite composition (Table~\ref{tab:paired}). Neither sampler
admits an ESS, an ELBO or a $\widehat Z_q$, because a controlled SDE has no
tractable density; and for the second, Corollary~\ref{cor:tilt} says a normalizer
comparison would report nothing missing even if a density were available. The
check returns $0.565$ against a true $0.559$ at $2.2\%$ of training cost and
clears its self-diagnostic.

\begin{table}[t]
\caption{Two training runs of the same controlled-SDE architecture on the same
target, with nearly equal missing mass and opposite decomposition. The
$\Msupp$ row is the audit of \S\ref{sec:exp} (five seeds); the $\Mwt$ row differs
only in the gradient clipping applied inside the control (two seeds). ESS, the
ELBO and $\widehat Z_q$ are undefined for both.}
\label{tab:paired}
\begin{center}
\begin{tabular}{lccccc}
\toprule
 & basins reached & $\Msupp$ & $\Mwt$ & $M$ & check: abs.\ error \\
\midrule
support collapse & 1/32  & 0.569 & 0.000 & 0.569 & 0.0025 \\
weight collapse  & 32/32 & 0.000 & 0.559 & 0.559 & 0.0057 \\
\bottomrule
\end{tabular}
\end{center}
\end{table}

\paragraph{The error floor at small $M$.} Weight collapse occupies a range of $M$
that the earlier experiments never visited, and there the estimator's own floor
matters. $\widehat M=\sum_j(\hat p_j-\hat q_j)_+$ rectifies rather than cancels
the noise in $\hat p_j$ and $\hat q_j$ at basins where $p_j\approx q_j$, so at
the population level $\mathbb E[\widehat M]\ge M$. On the GMM tilt family at a
true $M$ of zero the estimate reads $0.098$ at $R{=}256$ and $0.006$ at
$R{=}1024$; the mean absolute error over the family falls from $0.048$ to
$0.002$. We are careful about the direction of the effect: the mean signed error
at $R{=}256$ is positive ($+0.035$) as the rectification argument predicts, but
two of six rows come out low, so at this sample size the low-budget behaviour is
dominated by variance and we cannot claim a clean one-sided bias. The $M=0$ row
is the unambiguous measurement, since there non-negativity makes the whole
estimate error. On Many-Well the floor is already $0.007$ at $R{=}256$, so the
budget required is target-dependent. The practical consequence is that an
estimate small enough to be interesting -- say below $0.1$ -- should not be
trusted at a budget whose convergence has not been checked, which turns the
diagnostic of \S\ref{sec:selfcheck} from a useful guard into a requirement.

\section{A competitor that does not estimate a normalising constant}
\label{sec:altsearch}

The reference of \S\ref{sec:exp} estimates $\log Z$, and
Corollary~\ref{cor:tilt} bounds what that statistic can do: it is blind to
$\Mwt$ at the population level, so the weight-collapse comparison of
\S\ref{sec:weightcollapse} demonstrates a property of the estimand rather than a
contest. To make the comparison a contest we add a competitor that estimates the
basin weights \emph{directly} and never forms a normalising constant: parallel
tempering on a $16$-temperature ladder with replica exchange, reading
$\hat p_j$ off the occupancy of the $\beta=1$ replica, with no Hessian, no
Laplace approximation and no $Z$. It is charged in the same currency as
everything else.

\begin{table}[t]
\caption{A non-normalizer competitor on Many-Well ($K=32$, true $M=0.9854$).
Parallel tempering is not structurally blind the way the normalizer gap is: it
reaches the right answer given enough steps. It costs $15.7\times$ more to get
there. Costs are relative to \emph{this} sampler's deliberately short training
budget of $1.5\times10^{6}$ evaluations and are not comparable to the $2.7\%$
of Table~\ref{tab:main}, which is measured against the longer runs used there;
what is comparable is the ratio within this table.}
\label{tab:altsearch}
\begin{center}
\begin{tabular}{lccc}
\toprule
method & estimate & abs.\ error & cost \\
\midrule
ours & 0.9855 & 0.0001 & $55.7\%$ \\
parallel tempering, weights read directly ($10^3$ steps) & 0.9732 & 0.0122 & $291.7\%$ \\
parallel tempering, weights read directly ($3\times10^3$ steps) & 0.9854 & 0.0000 & $875.0\%$ \\
\bottomrule
\end{tabular}
\end{center}
\end{table}

Table~\ref{tab:altsearch} gives the comparison. Tempered MCMC is a genuine
alternative route to the missing mass -- unlike the normalizer gap it is not
blind to either term of \eqref{eq:decomp}, and at $3\times10^3$ steps it matches
the truth to four decimals. The difference is cost: it needs $15.7\times$ what
the check needs to get there, and at the cheaper setting it is still $0.012$ off
at $5.2\times$ the cost. Mode-search auditing is not the only way to obtain the
missing mass without ground truth; on these targets it is about an order of
magnitude cheaper than the tempered-MCMC route.

\section{Where the method stops: LJ-13}
\label{sec:scope}

Every target above has few, well-separated, non-degenerate modes. We now stress
the two load-bearing assumptions on the 13-atom Lennard-Jones cluster, whose
landscape has on the order of $1500$ local minima and whose global minimum is the
Mackay icosahedron at $E=-44.3268$ \citep{walesdoye1997}. We use the check's own
search machinery, not a better optimiser, because the question is whether the
check copes.

\paragraph{The search succeeds.} From random compact configurations, $98\%$ of
$6000$ restarts converge to $\|\nabla E\|<10^{-4}$, and all five independent
enumerations recover the global minimum exactly, $-44.3268$, matching the
literature value. They find between $413$ and $446$ distinct minima at $6000$
restarts, against a landscape holding on the order of $1500$.

\paragraph{Continuous symmetry breaks the local mass.} At the global minimum the
Hessian has exactly six zero eigenvalues -- three translations, three rotations --
with the next eigenvalue at $42.65$ and nothing in between. Equation
\eqref{eq:laplace} is therefore undefined on any target with continuous symmetry,
which includes essentially every molecular system. The repair is standard in
chemical physics: take the product over the $3N-6$ non-zero modes and treat the
zero modes through the rotational partition function and the point-group order,
as in the harmonic superposition approximation \citep{walesdoye1997}. With this
correction all $426$ usable minima admit a well-defined mass.

\paragraph{Enumeration is sufficient only at low temperature.} Relative harmonic
weights need only $\log w_j = -\beta E_j - \tfrac12\sum\log\lambda_j^{\neq0}$, so
one enumeration serves every $\beta$. Table~\ref{tab:lj} reports how much of the
weight a cheap search captures. At $\beta\ge8$ the weight sits entirely on the
global minimum, which the first $500$ restarts already find, and the check is
applicable. At $\beta\le3$ a $500$-restart search captures only $3$--$4\%$ of the
weight and the check must not be used. Two further points sharpen this. First,
the figures are optimistic: the denominator is the $6000$-restart set, not the
true $\sim\!1500$ minima. Second, and more damning for practice, the capture
fraction is not merely small but erratic -- at $\beta=1$ a $2000$-restart search
captures $23\pm26\%$, and even at $\beta=5$, where the mean looks usable, the
spread is $63\pm33\%$. A quantity with that variance cannot be acted on
regardless of where its mean sits, so we place the usable boundary at
$\beta\ge8$ rather than at the point where the mean first looks acceptable.

\begin{table}[t]
\caption{LJ-13. Concentration of the harmonic weight and the fraction of it
captured by a cheap search, as a function of inverse temperature. Mean $\pm$ s.d.\
over five independent enumerations of $6000$ restarts each. Note the standard
deviations in the last two columns: below $\beta=8$ the captured fraction is not
merely small but erratic.}
\label{tab:lj}
\centering
\begin{tabular}{rrrrrr}
\toprule
$\beta$ & heaviest mode & \#modes 90\% & \#modes 99\% & captured by 500 & by 2000 \\
\midrule
1  & $57.5\pm24.3\%$ & $6.8\pm3.3$ & $15.2\pm7.3$ & $3.0\pm2.9\%$   & $23.4\pm26.1\%$ \\
2  & $52.4\pm25.2\%$ & $7.2\pm3.5$ & $14.4\pm6.9$ & $3.5\pm2.9\%$   & $18.8\pm18.6\%$ \\
3  & $64.5\pm22.9\%$ & $5.0\pm2.9$ & $12.0\pm5.9$ & $3.7\pm4.0\%$   & $13.1\pm10.9\%$ \\
5  & $81.9\pm15.7\%$ & $2.0\pm0.6$ & $3.2\pm1.6$  & $63.3\pm33.3\%$ & $64.0\pm33.7\%$ \\
8  & $100.0\pm0.0\%$ & $1.0\pm0.0$ & $1.0\pm0.0$  & $100.0\pm0.0\%$ & $100.0\pm0.0\%$ \\
\bottomrule
\end{tabular}
\end{table}

\paragraph{There is no usable multimodal window on LJ-13.} A natural request at
this point is an end-to-end demonstration on LJ-13: train a sampler, let it
collapse, catch it. We cannot honestly supply one. At $\beta\ge8$, where the
enumeration is stable, the target is effectively unimodal -- the global minimum
holds all of the weight -- so there is no collapse to detect. At $\beta\le5$,
where several minima genuinely share the weight, the captured fraction is
$63\pm33\%$ or worse and the check must not be used. The two conditions do not
overlap on this system. Our end-to-end evidence with a modern sampler class is
therefore the controlled-SDE audit of \S\ref{sec:exp}, on Many-Well, which is one
of the standard targets of the neural-sampler literature
\citep{blessing2024beyond,chen2025scld,rissanen2025ptsd,guo2026pdns}; a
demonstration on a rugged physical landscape remains open.

\paragraph{A hypothesis we had to discard.} We expected the weight to concentrate
so quickly that enumerating the heavy modes would suffice regardless of
temperature. The middle columns of Table~\ref{tab:lj} appear to support this --
eight modes hold $90\%$ of the weight even at $\beta=1$ -- but the last two
columns refute it: a $500$-restart search captures only $3.2\%$. The heaviest
modes at high temperature are the entropically favoured, low-curvature ones, and
having a large harmonic weight is not the same as being easy for an optimiser to
fall into. We had implicitly assumed the two coincide. They do not.

\section{Related work}

\paragraph{The gap we are filling is stated by the evaluation literature itself.}
\citet{blessing2024beyond} build a standardised benchmark for sampling methods
and observe that model-support-restricted protocols are precisely the wrong
instrument for assessing mode collapse; \citet{wu2025rdsmc} record the same
problem operationally, noting a dependence on oracle metrics for hyperparameter
tuning. \citet{he2025notrick} supply the empirical phenomenon: simulation-free
neural samplers collapse without Langevin preconditioning, and do so while their
training objectives look healthy. Our contribution is the missing instrument, not
a new sampler.

\paragraph{Why generative-model coverage metrics do not transfer.} Evaluation of
generative models has converged on two-dimensional fidelity/coverage measures --
precision and recall \citep{sajjadi2018precision}, and its $k$-NN refinement
\citep{kynkaanniemi2019improved} -- precisely because scalar scores conflate
sample quality with mode coverage. These are \emph{two-sample} statistics: they
compare model draws against reference draws. In the neural-sampler setting there
are no reference draws, which is the whole difficulty; the same is true of
sample-based statistical distances more broadly. Our estimator is one-sample in
the target: it consumes $\tilde\pi$ and its derivatives, never samples of $\pi$.

\paragraph{Methods that fix collapse, rather than measure it.} A large recent
effort attacks the failure directly: learned Gaussian-mixture priors
\citep{blessing2025gmp}, affine-invariant tempering \citep{qin2025flowvat},
progressive tempering across temperatures \citep{rissanen2025ptsd}, proximal
schemes on path space \citep{guo2026pdns}, off-policy exploration with a novelty
bonus \citep{kim2025scalable}, and SMC-corrected transport
\citep{chen2025scld,wu2025rdsmc}. These are complementary to this paper in the
strict sense: each produces a sampler that our check can audit, and none of them
tells the practitioner whether the resulting run should be trusted. One family is
outside our reach rather than complementary to it: samplers over discrete state
spaces \citep{sanokowski2025discrete} admit no Hessian, so the local-mass step of
\S\ref{sec:method} has no analogue and the check does not apply. The closest
in spirit is \citet{dennehy2026weights}, who analyse when score-based models do
and do not recover mixture weights; they explain when the failure occurs, we
measure whether it has occurred in a given run.

\paragraph{What the check can be pointed at.} The audit needs draws from $q$ and
derivatives of $\tilde\pi$, and nothing else, so it is indifferent to how $q$ was
obtained. That covers the score-based variational families that have grown up
alongside diffusion samplers -- products of $t$-experts fitted by score matching
\citep{cai2026fisherfeynman}, low-rank score-based Gaussian families
\citep{modi2025bamp}, and their scaling to Bayesian neural networks
\citep{kim2026scorebnn} -- as well as flow families built for posterior geometry
\citep{ko2025mif}, regression-based surrogates trained on offline log-density
evaluations \citep{li2025nfr}, and diffusion-based variational posteriors
\citep{piriyakulkij2025ddvi,xu2026dbvi}. Two properties of these families make
the audit worth running: several are trained with mode-seeking objectives, and
several -- the diffusion-based ones in particular -- have no tractable density,
so the diagnostics of \S\ref{sec:blind} are not merely misleading there but
unavailable.

\paragraph{The estimator is imported, not invented.} Decomposing configuration
space into basins around local minima and summing local contributions is the
inherent-structure formalism of \citet{stillinger1984packing}, and summing
harmonic basin partition functions over enumerated minima is the harmonic
superposition approximation used throughout the energy-landscape literature
\citep{walesdoye1997,wales1998archetypal}. We take three things from that field
essentially unchanged: the enumeration by multi-start minimisation, the treatment
of the zero modes generated by continuous symmetry, and -- importantly -- its
known failure mode. The harmonic approximation to a basin free energy is
understood there to be reasonable at low temperature and to break down above a
system-dependent threshold, which is exactly the boundary our LJ-13 measurement
recovers from the outside (\S\ref{sec:scope}). Disconnectivity graphs
\citep{becker1997topology,wales1998archetypal} visualise the same decomposition.
What appears to be new here is the use: applying the construction not to predict a
physical partition function but to audit a learned sampler, and pairing it with a
statistic that reports when the enumeration underlying it is too incomplete to be
believed.

\section{Limitations}

The method rests on assumptions that our own experiments made load-bearing rather
than decorative. \textbf{(A1)} Modes must be reachable by multi-start
\emph{second-order} search; first-order search fails at condition number $20$,
which is mild by physical standards. \textbf{(A2)} Discovered modes must admit a
usable local mass; continuous symmetries require the zero-mode treatment of
\S\ref{sec:scope}, and modes whose anharmonicity differs by orders of magnitude
\emph{between} modes would break the cancellation that \eqref{eq:M} relies on.
\textbf{(A3)} The estimate is conditional on the enumerated landscape; under
incomplete enumeration its error has no fixed sign
(Proposition~\ref{prop:twosided}), which is what makes the accompanying
search-completeness diagnostic essential rather than decorative.
\textbf{(A4)} The cost claim is
dimension-limited: measured over $d=16$ to $128$ the check grows about linearly
in $d$ and would reach parity with training near $d\approx700$ in wall-clock, so
``a few percent'' describes present benchmark scales and not the largest problems
a neural sampler might be pointed at. All targets used here have exactly
computable ground truth, which was a deliberate choice and leaves behaviour on
rugged physical landscapes characterised only through the LJ-13 study.

\paragraph{The weight-collapse evidence is narrower than the rest.} Our exactly
constructed families cover two targets, not three: the tilted Many-Well sampler
is built coordinate-wise, which is correct only when the basin structure is
axis-aligned, and applying it to the rotated target produced a sampler that does
not match $\pi$ at all (its ESS is $10^{-4}$ where the construction should be
exact). We exclude those runs rather than report them. The paired controlled-SDE
result rests on two seeds of one architecture, and establishes that weight
collapse occurs, not how often.

\paragraph{The empirical split of $M$ is sample-size dependent.} We call a basin
unvisited when no draw of $q$ lands in it, so at $4\times10^4$ draws a basin
holding $10^{-5}$ of $q$'s mass is indistinguishable from an empty one. The
decomposition \eqref{eq:decomp} is exact as stated, but a basin visited with
vanishing probability falls under $\Mwt$ by our convention and under $\Msupp$
under a stricter one. Nothing in the estimator depends on the split; only the
attribution does.

\subsubsection*{Reproducibility statement}
All targets have closed-form or quadrature-exact ground truth, described in
\S\ref{sec:exp}. Every number in Tables~\ref{tab:selfcheck}--\ref{tab:lj} and
Figures~\ref{fig:selfcheck}--\ref{fig:frontier} is produced by the accompanying
code with fixed seeds.

\appendix

\section{Proofs}
\label{app:proofs}

\paragraph{Proposition~\ref{prop:ess}.} For $x\in S$ we have
$w(x)=\tilde\pi(x)/q(x)=Z\pi(x)\big/\big(\pi(x)/\omega\big)=Z\omega$, a constant
independent of $x$. Substituting $w_i\equiv Z\omega$ into
$\widehat{\mathrm{ESS}}=(\sum_i w_i)^2/(n\sum_i w_i^2)$ gives
$(nZ\omega)^2/(n\cdot n(Z\omega)^2)=1$. Since $q$ places no mass outside $S$ and
$S$ is a union of basins, $q_j=0$ for every basin outside $S$, so
$M=\sum_j(p_j-q_j)_+=\pi(S^c)=1-\omega$. \hfill$\square$

\paragraph{Proposition~\ref{prop:essdelta}.} Write $r=\log\{q/\pi(\cdot\mid S)\}$,
so $|r|\le\delta$ on $S$. For $x\in S$,
$\pi(x)=\pi(S)\,\pi(x\mid S)$ and hence
$w(x)=Z\pi(x)/q(x)=Z\,\pi(S)\,e^{-r(x)}$, which lies in
$[ae^{-\delta},ae^{\delta}]$ with $a=Z\pi(S)$. Therefore
$(\mathbb E_q w)^2\ge a^2e^{-2\delta}$ and $\mathbb E_q[w^2]\le a^2e^{2\delta}$,
so $\mathrm{ESS}=(\mathbb E_q w)^2/\mathbb E_q[w^2]\ge e^{-4\delta}$. The bound
contains no reference to $\pi(S)$, which cancels, and hence none to
$\Msupp=1-\pi(S)$. \hfill$\square$

\paragraph{Proposition~\ref{prop:identity}.} Since $q$ vanishes off $S$,
$\mathbb E_q[\tilde\pi/q]=\int_S (\tilde\pi/q)\,q=\int_S\tilde\pi=Z\int_S\pi
=Z\pi(S)$. As $S$ is a union of basins, the basins outside $S$ are exactly those
with $q_j=0$, so $\pi(S)=1-\sum_{j:q_j=0}p_j=1-\Msupp$. \hfill$\square$

\paragraph{Corollary~\ref{cor:tilt}.} If
$\operatorname{supp}(q)\supseteq\operatorname{supp}(\pi)$ then no basin has
$q_j=0$, so $\Msupp=0$ and Proposition~\ref{prop:identity} gives
$\mathbb E_q[\tilde\pi/q]=Z$ irrespective of the values $q_j$. Hence the
population value of $1-\widehat Z_q/\widehat Z_{\mathrm{ref}}$ is $1-Z/Z=0$. The
bound on $\Mwt$ is attained by putting all of $q$'s mass on the heaviest basin
in the limit, giving $\sum_{j\ne j^\star}p_j=1-\max_j p_j$. \hfill$\square$

\paragraph{Corollary~\ref{cor:silent}.} Immediate:
$1-\widehat Z_q/\widehat Z_{\mathrm{ref}}\le 0$ iff
$\widehat Z_q\ge\widehat Z_{\mathrm{ref}}$. \hfill$\square$

\paragraph{Proposition~\ref{prop:nocert}.} Let $x_1,\dots,x_N$ be the points the
algorithm queries when run on $\tilde\pi_1$, and let $Z_1=\int\tilde\pi_1$. Put
$\tilde\pi_2=\tilde\pi_1+\lambda\varphi_z$ with
$\varphi_z(x)=(2\pi)^{-d/2}e^{-\|x-z\|^2/2}$ and $\lambda=Z_1(1-\delta)/\delta$,
so that the bump carries a fraction
$\lambda/(Z_1+\lambda)=1-\delta$ of the total mass of $\tilde\pi_2$ and therefore
$M(\tilde\pi_2)\ge1-\delta$, while $M(\tilde\pi_1)$ may be $0$.

Write $\rho=\|x-z\|$. Direct differentiation gives
$\varphi_z(x)=(2\pi)^{-d/2}e^{-\rho^2/2}$,
$\|\nabla\varphi_z(x)\|=\rho\,\varphi_z(x)$ and
$\|\nabla^2\varphi_z(x)\|_{\mathrm{op}}\le(\rho^2+1)\varphi_z(x)$, so all three
perturbations are bounded by $(1+\rho+\rho^2)\varphi_z(x)$. For $\rho\ge4$ we have
$3\rho^2\le e^{\rho^2/4}$ and hence
$(1+\rho+\rho^2)e^{-\rho^2/2}\le3\rho^2e^{-\rho^2/2}\le e^{-\rho^2/4}$.
Therefore, if $z$ is at distance at least $r$ from every queried point and $r$
satisfies \eqref{eq:bumpr}, every queried value and derivative changes by at most
$\lambda(2\pi)^{-d/2}e^{-r^2/4}\le\varepsilon$. Reading at precision
$\varepsilon$, the algorithm cannot distinguish the two targets: it follows the
same execution path and returns the same estimate, which therefore errs by at
least $(1-\delta)/2$ on one of them.

Such a $z$ exists: the $N$ excluded balls of radius $r$ occupy volume
$N\,V_d r^d$ with $V_d$ the unit-ball volume, so any ball of radius
$R_0>r\,N^{1/d}$ has volume $V_dR_0^d>N V_d r^d$ and cannot be covered by them.
\hfill$\square$

\paragraph{Proposition~\ref{prop:twosided}.} Both cases are direct computation.
For the first, $\sum_{k\in S}p_k=0.7$ gives $\hat p=(0.4,0.3)/0.7=(0.571,0.429)$;
assigning the third basin's $q$-mass $0.4$ to the second discovered mode gives
$\hat q=(0.2,0.4+0.4)=(0.2,0.8)$; hence
$\widehat M=(0.571-0.2)_++(0.429-0.8)_+=0.371$, while
$M=(0.4-0.2)_++(0.3-0.4)_++(0.3-0.4)_+=0.2$. For the second, $S$ contains one
mode with $p_1=q_1$ after normalisation, so every term vanishes and
$\widehat M=0$, while the undiscovered mode contributes $p_2=0.5$ to $M$.
\hfill$\square$

\paragraph{Proposition~\ref{prop:bias}.} Write $M_{\mathrm{tot}}=\sum_k m_k$.
From $\hat p_j=c_jm_j/\sum_k c_km_k$ and the assumed bounds,
$\hat p_j\le \frac{c(1+\eta)m_j}{c(1-\eta)M_{\mathrm{tot}}}=p_j\frac{1+\eta}{1-\eta}$
and symmetrically $\hat p_j\ge p_j\frac{1-\eta}{1+\eta}$, so
$|\hat p_j-p_j|\le p_j\frac{2\eta}{1-\eta}$. Since $t\mapsto(t)_+$ is
$1$-Lipschitz,
$|\widehat M-M|\le\sum_j|\hat p_j-p_j|\le\frac{2\eta}{1-\eta}\sum_j p_j
=\frac{2\eta}{1-\eta}$. \hfill$\square$

\paragraph{Proposition~\ref{prop:blind}.} Both statistics are functions of the
set of modes discovered at the two budgets. A mode reached by no restart appears
in neither set, hence changes neither the numerator nor the denominator of either
ratio, and the verdict is the same as if that mode did not exist. The restarts
being independent, the probability that mode $j$ is reached by none of the $R$ is
$(1-\alpha_j)^{R}$, which is bounded away from zero whenever $\alpha_j R=O(1)$.
\hfill$\square$

\bibliographystyle{plainnat}
\bibliography{references}

\end{document}